\documentclass[conference,letter]{IEEEtran}
\IEEEoverridecommandlockouts

\usepackage{amsmath,amssymb,mathtools}
\usepackage{bm}
\usepackage{tabularx}
\usepackage{booktabs}
\usepackage{xcolor}
\usepackage{soul}

\usepackage{graphicx} % Required for including images
\usepackage{caption}
\usepackage{subcaption}
\graphicspath{{figures/}} 
\usepackage{enumitem}
\setlist[enumerate]{leftmargin=*}
\setlist[itemize]{leftmargin=*}

\usepackage[autostyle]{csquotes}
\MakeOuterQuote{"}
\usepackage{hyperref}
\hypersetup{
    pdfauthor={},
    pdftitle={},
    pdfsubject={},
    pdfkeywords={},
    colorlinks=false, % IEEE style prefers boxed links
    pdfborder={0 0 0}
}

\usepackage{algorithm}
\usepackage{algpseudocode}

\usepackage[
    style=numeric,%ieee, 
    sorting=none,        % Bibliography is ordered by first appearance in the text
    sortcites=true,      % Sorts numbers numerically INSIDE the \cite{} brackets
    backend=biber, 
    maxbibnames=99,      % Forces all authors to be listed in the bibliography
    giveninits=true,     % Abbreviates first names to initials (e.g., John Doe -> J. Doe)
    natbib=true,
    doi=false,   % Excludes DOI fields
    url=false,   % Excludes URL fields
    isbn=false,  % Optional: Excludes ISBN (common for IEEE to save space)
    eprint=false % Optional: Excludes eprint/arXiv links    
]{biblatex}

\usepackage{amsthm}

\newtheorem{proposition}{Proposition}
\theoremstyle{definition}
\newtheorem{definition}{Definition}
\theoremstyle{remark}

\newcommand{\mytitle}{\LARGE \bf  Muscle Coactivation in the Sky: Geometry and\\ Pareto Optimality of Energy  vs.  Aerodynamic Promptness\\ and Multirotors as Variable Stiffness Actuators}

\title{\mytitle \vspace{-0.5em}}

\author{%
Antonio Franchi$^{1,2}$%
\thanks{$^1$ Robotics and Mechatronics Department, Electrical Engineering, Mathematics, and Computer Science (EEMCS) Faculty, University of Twente, 7500 AE Enschede, The Netherlands. {\footnotesize schol@r-franchi.eu}}
\thanks{$^2$Department of Computer, Control and Management Engineering, Sapienza University of Rome, 00185 Rome, Italy. {\footnotesize schol@r-franchi.eu}}
\thanks{This work was partially funded by the Horizon Europe research agreement no. 101120732 (AUTOASSESS).}
\vspace{-3em}
}

\newif\ifarxiv
\arxivtrue  % Change to \arxivfalse for the submission version
\begin{document}
\maketitle
\begin{abstract}
In robotics and biomechanics, trading metabolic cost for kinematic readiness is a well-established principle. This paper formalizes this concept for aerial multirotors through the introduction of aerodynamic promptness---a dynamic metric analogous to dynamic manipulability in robotics. By formulating redundancy resolution as a geometric multi-objective optimization along task fibers, we rigorously characterize the topological trade-off between energy consumption and promptness. We demonstrate that this interplay is fundamentally governed by fiber geometry. Cooperative actuation regime yields compact fibers with bounded, compatible Pareto fronts. Conversely, antagonistic actuation regime unlocks unbounded fibers, enabling  aerodynamic co-contraction that drives promptness to hardware limits at the expense of flight endurance. We establish a structural isomorphism between  aerodynamic co-contraction and biologically inspired variable stiffness actuators, introducing a dynamic ``flying muscle'' paradigm. Ultimately, this framework transitions multirotor allocation from heuristic energy minimization to principled, geometry-aware Pareto navigation, laying foundational theory for the design and control of highly agile aerial platforms.
\end{abstract}

%%% %%%%%%%%%%%%%%%%%%
\section{Introduction}
\label{sec:introduction}
%%% %%%%%%%%%%%%%%%%%%

Multirotor vehicles with actuation redundancy---including standard hexarotors, fully actuated platforms~\cite{2018e-TogFra,Brescianini2016ICRA,Welde-2024-Riemannian}, and servomotor-equipped designs~\cite{Ryll2015TCST,bodie2020_iser_volirox}---possess a continuous null-space of admissible actuator commands. Standard allocation strategies typically resolve this freedom by minimizing control effort via pseudo-inverses or quadratic programming~\cite{Bodson2002JGCD,Ryll2015TCST,Brescianini2016ICRA,park2018_tmech_odar,bodie2020_iser_volirox}. Contemporary autopilots apply these principles to handle actuator bounds and failures~\cite{PX4ControlAllocation,ArduPilotMixer}. However, energetically minimal allocations often bias the system toward ``lazy'' solutions, leaving the platform kinematically sluggish near singular configurations.

A multirotor's ability to reject strong disturbances~\cite{Romero2022TRO_MPCC} or execute agile maneuvers~\cite{kaufmann2020_rss_acrobatics,saviolo2022_ral_pitcn} depends critically on its \emph{promptness}: the capacity to produce rapid wrench variations under physical rotor acceleration limits. This concept mirrors kinematic manipulability~\cite{Yoshikawa1985IJRR}. While manipulability in aerial robotics is typically reserved for attached manipulators~\cite{Ruggiero2018RAL} or static design optimization~\cite{Hamandi2021IJRR,arza2025_taskopt}, applying it dynamically to the rotor-speed configuration space exposes a fundamental conflict between flight endurance and dynamic readiness.
In biomechanics, trading metabolic cost for kinematic readiness via structural co-contraction is a well-established principle~\cite{Burdet2001Nature,Franklin2003JNP,Gribble2003JNP,Gogeascoechea-2024}, artificially echoed in robotics through impedance control, energy-tank protocols, or compliant actuators~\cite{Hogan1984ACC,Califano2022RAL_EnergyTanks,Amara2021RAL_FieldWeakening,Wolf-vsa-2016}. Yet, this dynamic paradigm has never been formally characterized for multirotor-based aerial robots. The main thesis of this work is that the trade-off between efficiency and promptness is not merely heuristic, but deeply rooted in the geometric topology of the actuation task space.

\emph{Contributions:}
We bridge biomechanics and flight control by formalizing aerodynamic co-contraction, demonstrating that redundant multirotors can function mathematically as a ``flying muscle'' group.
\begin{enumerate}
    \item \emph{Geometric Promptness Formulation:} We reinterpret manipulability as a  measure of fiber density on the rotor-speed manifold. We show that antagonistic rotor regime structurally enlarges the available wrench-rate volume.
    \item \emph{Topological Trade-off Characterization:} We rigorously define the conflict between aerodynamic power (an $L_3$ norm on rotor speeds) and promptness. We prove this interplay is strictly governed by task fiber geometry: compact (cooperative) fibers yield bounded trade-offs, whereas unbounded (antagonistic) fibers induce severe, fundamental conflict.
    \item \emph{The ``Flying Muscle'' Isomorphism:} We establish a mathematical equivalency between aerodynamic co-contraction and biological variable stiffness actuators, offering a principled theoretical foundation for tuning aerial compliance beyond standard minimum-energy allocators.
\end{enumerate}

\smallskip
\subsubsection*{Paper Organization}
% The paper is structured as follows. 
Sections~\ref{sec:fiber_optimization} and~\ref{sec:moo_fibers} establish the geometric framework for fiber-based multi-objective optimization, framing the operational questions in Section~\ref{sec:fundamental_questions}. Section~\ref{sec:framework} adapts this to multirotors, defining aerodynamic promptness. Section~\ref{sec:case_study} analyzes the compatible-to-antagonistic topological shift via a dual-rotor case study. Section~\ref{sec:vsa_connection} details the isomorphism to biological variable stiffness actuators, and Section~\ref{sec:conclusion} concludes the paper.

%%% %%%%%%%%%%%%%%%%%%
\section{Math Preliminary 1: Optimization of Fiber Density in Nonlinear Maps}
\label{sec:fiber_optimization}
%%% %%%%%%%%%%%%%%%%%%

Let $\mathcal{V}$ be a smooth manifold of dimension $n$, representing a \emph{total (configuration) space}, endowed with a Riemannian metric $g$ having the local coordinate matrix representation $G_v\in S^n_{++}$ for each $v\in \mathcal{V}$. In the robotics context, the space $\mathcal{V}$ may represent, e.g., the configuration of a robot or of a group of actuators. 
Let $\mathcal{W}$ be a \emph{base (task) space} of dimension $m < n$. 
A surjective submersion $f:\mathcal{V}\to\mathcal{W}$ induces a natural equivalence relation on $\mathcal{V}$, partitioning the total space into a disjoint union of closed submanifolds known as fibers. 
\begin{definition}[Fiber]
Given a task $w \in \mathcal{W}$, the \emph{fiber} associated with $w$ is the level set:
\begin{equation}
    \mathcal{F}_w \;\coloneqq\; f^{-1}(w) \;=\; \{\,v \in \mathcal{V} \mid f(v) = w\,\}.
\end{equation}
By the implicit function theorem, $\mathcal{F}_w$ is a smooth submanifold of $\mathcal{V}$ of dimension $n - m$.
\end{definition}
Geometrically, this structures $\mathcal{V}$ as a fiber bundle where $\mathcal{V} = \bigsqcup_{w \in \mathcal{W}} f^{-1}(w)$, with each target state $w \in \mathcal{W}$ serving as the representative element for its corresponding equivalence class.

At any  $v \in \mathcal{V}$, the differential $df_v: T_v\mathcal{V} \to T_{f(v)}\mathcal{W}$ induces a direct sum decomposition of the tangent space:
$$T_v\mathcal{V} = V_v \oplus H_v.$$
The vertical space $V_v = \ker(df_v)$ is completely determined by the fibers, containing all motions that leave the base space coordinate $w$ invariant. The horizontal space $H_v = V_v^\perp$ is the orthogonal complement of the vertical space, governed simultaneously by the topological structure of the fibers and by the metric $g$, which defines the orthogonality condition.

Given a desired \emph{task rate} (a tangent vector) $dw \in T_w\mathcal{W}$ at a base point $w = f(v)$, the surjectivity of $df_v$ implies the existence of an infinite set of compatible \emph{configuration rates} $dv \in T_v\mathcal{V}$. However, to realize the specified $dw$ locally while exerting the minimum effort according to the metric $g$, the motion must be restricted entirely to the horizontal space $H_v$. Restricted to this orthogonal subspace, the linear mapping $df_v|_{H_v} : H_v \to T_w\mathcal{W}$ is an isomorphism, providing a unique, optimal local realization of the prescribed tangent vector $dw$.

While the horizontal space provides the optimal instantaneous motion $dv$ for a fixed $v\in f^{-1}(w)$, a higher-level redundancy resolution problem remains: selecting the optimal configuration $v$ within the fiber $f^{-1}(w)$. To minimize the expected norm of the rate in $T_v\mathcal{V}$ required to generate a generic basis in $T_{f(v)}\mathcal{W}$, a geometrically sound objective is to select a point $v$ along the fiber that maximizes the density of adjacent fibers in its local neighborhood. At such dense configurations, a minimal displacement in $T_v\mathcal{V}$ (more precisely in $H_v$) yields a maximal variation in $T_{f(v)}\mathcal{W}$. 

This orthogonal fiber density is fundamentally dependent on the metric $g$ in two distinct capacities: first, to establish the orthogonal projection onto $H_v$, and second, to quantify the density of the fibers restricted to this subspace or, equivalently,  the aforementioned `expected' norm.
The mathematical framework describing this spatial density is governed by the Coarea Formula. For a measurable function $\phi$ on $\mathcal{V}$, the integration over the $\mathcal{V}$ relates to the integration over the fibers via:
$$\int_{\mathcal{V}} \phi(v) J_f^N(v) dv = \int_{\mathcal{W}} \left( \int_{f^{-1}(w)} \phi(v) d\sigma \right) dw,$$
where $J_f^N(v)$ is the normal Jacobian, serving as the exact measure of orthogonal fiber density.

In robotics, this geometric density is structurally equivalent to the concept of dynamic manipulability \cite{Yoshikawa1985Dynamic}. The determinant of the normal Jacobian is proportional to the volume of the ellipsoid in $T_{f(v)}\mathcal{W}$ associated with a pushed-forward Riemmannian metric $g$. In local coordinates, let $J_f(v)$ be the Jacobian of $f$ at $v$ representing $df_v$ in coordinates. The metric in $T_{f(v)}\mathcal{W}$ is constructed by pushing forward the quadratic form of the co-metric $g^{-1}$  at $v$, yielding:
\begin{align}
D^{-1}(v) = (J_f(v) G^{-1}_v J_f^T(v))^{-1}.
    \label{eq:pushed_forward_metric}
\end{align}
This expression dictates the composition and subsequent inversion of three sequential linear maps: the adjoint of the differential $df_v^*$, represented by $J_f^T(v)$; the co-metric induced by $g$, represented by $G^{-1}_v$; and the differential $df_v$, represented by $J_f(v)$ itself. To obtain a quantity proportional to the orthogonal fiber density at $v$, one evaluates the determinant of the pushed-forward co-metric $D(v) = J_f(v) G^{-1}_v J_f^T(v)$:
\begin{align}
    \rho(v) = \sqrt{\det(D(v))},
    \label{eq:fiber_density}
\end{align}
which is the generalization of the familiar formula of dynamic manipulability in robotics to the fiber density of any generic surjective submersion $f:\mathcal{V}\to\mathcal{W}$. Given the multiple interpretations and historical connections, we shall use the terms (dynamic) manipulability, fiber density, and (kinematic) \emph{promptness}\footnote{The word ``promptness'' here must be intended as the ability to produce larger task rates with configuration rate norms remaining equal.} equivalently to denote the same scalar map $\rho$. 

\begin{definition}[density/promptness  optimization]
The global redundancy resolution problem is then formulated as the maximization of the density $\rho(v)$ constrained to the fiber:
\begin{align}
\label{eq:density_maximization}
\text{given } w\in \mathcal{W}: \quad \max_{v \in \mathcal{V}} \rho(v) \quad \text{subject to} \quad f(v) = w.
\end{align}
\end{definition}

\subsection{Invariance of the Solution Landscape}

It is important to verify that the optimal solutions to \eqref{eq:density_maximization} are intrinsically invariant to arbitrary local coordinate choices on both $\mathcal{V}$ and $\mathcal{W}$, as well as to any specific Riemannian metric imposed on $\mathcal{W}$.
Under a change of basis defined by Jacobians $T_B$ on $\mathcal{V}$ and $T_C$ on $\mathcal{W}$, the pushforward matrix transforms as $\tilde{D} = T_C J_f T_B (T_B^T G T_B)^{-1} T_B^T J_f^T T_C^T$. The dependency on the domain coordinates cancels, yielding $\tilde{D} = T_C D T_C^T$. Thus, the core pushforward expression is completely independent of the coordinates chosen in $\mathcal{V}$. 

To account for an arbitrary Riemannian metric $h$ on the task space $\mathcal{W}$, having a local matrix representation $H$, the volume of the pushforward ellipsoid is generalized to incorporate this metric as $\rho_H(v) = \sqrt{\det(D(v)H)}$. The influence of both the target coordinates and this chosen target metric is strictly punctual in $\mathcal{W}$. Because the optimization search operates exclusively along a single fiber, the base task $w = f(v)$ remains geometrically fixed. Factoring the determinant yields:
$$\rho_H(v) = \sqrt{\det(H(w))} \sqrt{\det(D(v))}.$$
Similarly, under a change of target coordinates, the unweighted volumetric measure scales as $\sqrt{\det(\tilde{D})} = |\det(T_C(w))| \sqrt{\det(D(v))}$. Consequently, both the metric determinant $\sqrt{\det(H(w))}$ and the coordinate transformation determinant $|\det(T_C(w))|$ evaluate at the constant base task $w$, behaving exclusively as positive multiplicative constants in the reward of~\eqref{eq:density_maximization}. By the elementary properties of the determinant operator and continuous optimization, uniformly scaling an objective function by a positive constant preserves the exact locations of its extrema. While the reward functions might deform or shift under such mappings across different fibers, the geometric density defined in this manner guarantees an invariant solution landscape of \eqref{eq:density_maximization} within each single fiber.

\subsection{Discussion and Summary}

In robotics applications, the maximization of fiber density serves to minimize the expected configuration-space rate $dv$ (e.g.,  joint velocities) to generate task-space variations $dw$ needed, e.g., for control. This is highly relevant in the presence of limitations/saturations on $dv$ imposed by the physical system. Positioning the system on a $v$ by solving \eqref{eq:density_maximization} may guarantee stability and performance otherwise impossible in other configurations.

The converse objective holds equal practical significance. In operational scenarios prioritizing high precision or requiring resilience to actuation noise within the configuration space $\mathcal{V}$, one may instead seek to minimize the orthogonal fiber density. Operating in regions of minimal density ensures that bounded disturbances or control inaccuracies in the configuration space are geometrically attenuated when mapped through the differential to the base space $\mathcal{W}$. From an optimization perspective, transitioning from density maximization to minimization simply inverts the objective function; the fundamental invariance properties previously established remain entirely unaffected.

\subsubsection*{Summary} the existence of a nonlinear surjective submersion $f: \mathcal{V} \to \mathcal{W}$ alongside a Riemannian metric $g$ on the total space $\mathcal{V}$ intrinsically induces a global scalar field $\rho:\mathcal{V}\to \mathbb{R}_{\geq 0}$, which quantifies the orthogonal fiber density restricted to the horizontal subspace at any configuration $v$ and acts as a measure of kinematic promptness, also known as dynamic manipulability in robotics. Mathematically anchored by the normal Jacobian in the coarea formula and physically manifested as the dynamic manipulability measure in kinematics, this density is proportional to the volume of the ellipsoid generated by pushing forward to $\mathcal{W}$ the metric $g$ via the differential $df$ and its adjoint. Whether the operational objective necessitates maximizing or minimizing this density, the resulting constrained optimization problem evaluated along the fibers is geometrically well-posed. The optimal configurations for a given task $w\in\mathcal{W}$ are governed strictly by the intrinsic geometry of $(\mathcal{V}, g)$ and the map $f$, rendering them  invariant to the choice of local coordinates on both $\mathcal{V}$ and $\mathcal{W}$, as well as to any auxiliary metric assigned to the task space $\mathcal{W}$.

%%% %%%%%%%%%%%%%%%%%%%
\section{Math Preliminary 2: Multi-Objective Optimization Alongside Density on Fibers}
\label{sec:moo_fibers}
%%% %%%%%%%%%%%%%%%%%%%

Having established the intrinsic geometric properties of the orthogonal fiber density $\rho(v)$---induced implicitly by a nonlinear map $f:\mathcal{V}\to\mathcal{W}$ and the choice of a Riemannian metric $g$ in $\mathcal{V}$---we now introduce an additional scalar field $h: \mathcal{V} \to \mathbb{R}$ representing a configuration-dependent cost, such as the energetic or power-related effort required to maintain or actuate the system at a specific configuration $v \in \mathcal{V}$. Physical costs of this nature are generally independent of both the nonlinear kinematic map $f$ and the density field $\rho(v)$. 

Furthermore, physical systems are subject to hardware limitations (e.g., actuator saturations and joint limits), which define a restricted domain of feasibility $\overline{\mathcal{V}} \subset \mathcal{V}$. 

The core analytical challenge thus becomes characterizing the interplay between these two competing scalar objectives---kinematic fiber density $\rho$ and energetic cost $h$---strictly within the feasible regions of the configuration space that satisfy a given task, i.e.,   $f^{-1}(w)\cap\overline{\mathcal{V}} $ for a given $w\in \mathcal{W} $. We formalize this dual-objective problem into a general geometric setting for multi-objective optimization on the fibers. This formalization details where the redundancy lies, how objectives interact when restricted to a redundancy manifold, and how physical inequalities reshape this topological interaction.

\begin{definition}[Riemannian Projector]
Using the Riemannian metric $g$ on $\mathcal{V}$, the $g$-orthogonal projector onto the vertical space, $P_v: T_v\mathcal{V} \to V_v$, is given by:
\begin{equation}
    P_v \;=\; I_n - G^{-1}_v J_f^{T}(v) \Big(J_f(v) G^{-1}_v J_f^{T}(v)\Big)^{-1} J_f(v).
\end{equation}
This operator isolates the internal null-space motions that do not alter the task $f(v)$, properly weighted by the metric $g$.
\end{definition}

Let $J_1, J_2: \mathcal{V} \to \mathbb{R}$ represent the two competing smooth objectives. In the current context, $J_1(v) = h(v)$ models the energetic cost, and $J_2(v) \propto -\rho^2(v) = -\det(D(v))$ maps the maximization of fiber density into a minimization problem. 

\begin{definition}[Restricted Gradient on the Fiber]
Let $\nabla J_i(v)$ denote the standard Euclidean gradient of $J_i(v)$ for $i\in\{1,2\}$. The Riemannian gradient is $\text{grad}_g J_i(v) = G^{-1}_v\nabla J_i(v)$. The gradient of $J_i$ strictly restricted to the fiber $\mathcal{F}_w$ is obtained via orthogonal projection:
\begin{equation}
    \nabla_{\mathcal{F}_w}J_i(v) \;\coloneqq\; P_v\,\text{grad}_g J_i(v) \,\in\, V_v.
\end{equation}
A configuration $v$ is termed a \emph{fiber-stationary} point for $J_i$ when $\nabla_{\mathcal{F}_w}J_i(v) = 0$.
\end{definition}

\begin{definition}[Local Alignment Index]
Let $v \in \mathcal{F}_w$ where $\nabla_{\mathcal{F}_w}J_1(v)$ and $\nabla_{\mathcal{F}_w}J_2(v)$ are not both zero. The local conflict between the objectives is quantified by:
\begin{equation}
    \kappa(v) \;=\; \frac{g_v\Big(-l_1(v),\, -l_2(v)\Big)}
    {\sqrt{g_v\Big(l_1(v), l_1(v)\Big)} \sqrt{g_v\Big(l_2(v), l_2(v)\Big)}}\in [-1, 1],
\end{equation}
where we use the shorthand $l_i \coloneqq \nabla_{\mathcal{F}_w}J_i(v)$. 
The regimes are classified as follows:
\begin{itemize}
    \item $\kappa(v) = 1$: the restricted descent directions are aligned (local \emph{compatibility}).
    \item $\kappa(v) = -1$: the restricted descent directions are anti-aligned (local \emph{conflict}).
    \item $\kappa(v) = 0$: the restricted descent directions are orthogonal (local \emph{independence}).
\end{itemize}
If both restricted gradients vanish simultaneously, $\kappa(v)$ is undefined, though functionally this represents a shared fiber-stationary point (ideal compatibility).
\end{definition}

\begin{definition}[Local Pareto Optimality on the Fiber]
A configuration $v^* \in \mathcal{F}_w$ is a local Pareto optimum if there exists no non-zero admissible displacement $\delta v \in V_{v^*}$ that strictly decreases both objectives:
$$g_v\Big(\nabla_{\mathcal{F}_w}J_1(v^*),\, \delta v\Big) < 0 \quad \text{and} \quad g_v\Big(\nabla_{\mathcal{F}_w}J_2(v^*),\, \delta v\Big) < 0.$$
Equivalently, there must exist scalars $\lambda_1, \lambda_2 \ge 0$, not both zero, such that:
\begin{equation}
    \lambda_1\,\nabla_{\mathcal{F}_w}J_1(v^*) \;+\; \lambda_2\,\nabla_{\mathcal{F}_w}J_2(v^*) \;=\; 0.
\end{equation}
For strictly non-zero gradients, this condition reduces to perfect anti-parallelism ($\kappa(v^*) = -1$).
\end{definition}

\smallskip
Considering the aforementioned hardware limitations and feasibility set, the relevant optimization domain is therefore the \emph{feasible fiber}:
\begin{equation}
    \tilde{\mathcal{F}}_w \;\coloneqq\; \mathcal{F}_w \cap \overline{\mathcal{V}}.
\end{equation}
In the interior of $\tilde{\mathcal{F}}_w$, the previously defined tangent-space conditions hold exactly. At boundary configurations, the vertical space $V_v$ is replaced by the intersection of $V_v$ and the local tangent cone of $\overline{\mathcal{V}}$, with Pareto conditions reformulated over feasible restricted descent directions.

\begin{definition}[Pareto Set and Conflict Measure]
Let $\mathcal{P}_w \subset \tilde{\mathcal{F}}_w$ define the set of local Pareto optima on the feasible fiber. The mapped image $\mathbf{J}(\mathcal{P}_w) = \{(J_1(v), J_2(v)) \mid v \in \mathcal{P}_w\} \subset \mathbb{R}^2$ forms the Pareto front in the objective space. A global index of objective antagonism is the \emph{Pareto extent}, defined by the geometric arc length of $\mathbf{J}(\mathcal{P}_w)$. An extent approaching zero signifies high global compatibility, whereas a large or unbounded extent indicates a severe performance trade-off.
\end{definition}

The interaction between energetic cost and kinematic density is critically \emph{fiber-dependent}. As the task $w$ changes in $\mathcal{W}$, or as the feasibility subset $\overline{\mathcal{V}}$ bounds different regions of the full fiber, the topology of $\tilde{\mathcal{F}}_w$ can undergo fundamental transformations (e.g., from compact to non-compact sets). 

For example, consider an abstract task constraint $v_1|v_1| + v_2|v_2| = w$ with $w > 0$ (see Sec.~\ref{sec:case_study}). If $\overline{\mathcal{V}}$ restricts operations to the first quadrant ($v_1, v_2 > 0$), the feasible fiber is a compact circular arc $v_1^2 + v_2^2 = w$. Continuous objectives evaluated along this compact arc naturally attain finite minimizers in close proximity. Conversely, if $\overline{\mathcal{V}}$ restricts operations to $v_1 < 0, v_2 > 0$, the feasible fiber transforms into a non-compact hyperbolic branch $-v_1^2 + v_2^2 = w$. Antagonistic objective gradients may now drive solutions toward infinity along diverging branches, creating a fundamentally different operational regime. Thus, whether density and energy are cooperative or conflicting is not an absolute property of the scalar fields themselves, but depends inherently on the geometry of the specific feasible fiber $\tilde{\mathcal{F}}_w$ being traversed when restricted to $\overline{\mathcal{V}}$.

%%% %%%%%%%%%%%%%%%%%%%%%%%
\section{Fundamental Research Questions in the Tradeoff between Energy Cost and Fiber Density}
\label{sec:fundamental_questions}
%%% %%%%%%%%%%%%%%%%%%%%%%%

Structuring redundancy resolution through this geometric lens motivates three guiding questions for operational analysis:
\begin{enumerate}
    \item \emph{Compatibility along a specific fiber:} For a fixed task $w$, how do the energy and density objectives interact on $\tilde{\mathcal{F}}_w$? Do they share an unconstrained minimizer, or does their restricted gradient alignment $\kappa(v)$ reveal severe local conflict?
    \item \emph{Topological dependence across the task space:} Does the objective interplay remain qualitatively consistent across $\mathcal{W}$, or do geometric variations in $\mathcal{F}_w$ (such as transitions in compactness or severe metric conditioning) trigger regime switches between compatibility and conflict?
    \item \emph{Impact of hardware constraints:} How do strict physical inequalities in $\overline{\mathcal{V}}$ truncate the Pareto set $\mathcal{P}_w$? Do boundary conditions neutralize conflicts that exist in the unbounded total space, or do they artificially induce them by restricting access to shared fiber-stationary points?
\end{enumerate}

%%% %%%%%%%%%%%%%%%%%%%%
\section{A General Framework for Energy vs. Aerodynamic Promptness in Multirotors}
\label{sec:framework}
%%% %%%%%%%%%%%%%%%%%%%%

We now specialize the general geometric multi-objective framework to the control allocation of multirotor UAVs. The core contribution of this part is demonstrating that the nonlinear mapping from propeller rotational speeds to the generated body wrench, coupled with the physical rate limits on rotor acceleration, naturally necessitates the optimization of  density/promptness. Without these two physical ingredients, promptness as an objective would be trivial: it would be physically meaningless under infinite acceleration capabilities, and strictly constant under a purely linear force model. By formally establishing this promptness as a dynamic, state-dependent measure of fiber density, we unlock the ability to rigorously evaluate its trade-off against energy consumption along the feasible fibers of the system. 

By adopting standard, universally accepted multirotor actuation models for this analysis, we demonstrate that the dichotomy between energy efficiency and control authority is not a fragile artifact of complex aerodynamic assumptions, but a fundamental, first-order characteristic emerging directly from the basic physics of flight.

\smallskip
\subsubsection*{The Nonlinear Direct Kinematic Map}

Let the total configuration manifold $\mathcal{V} \subset \mathbb{R}^n$ represent the space of propeller rotational speeds, where the coordinate $v_i$ dictates the angular rate of the $i$-th brushless motor. We note that $\mathcal{V}$ is a proper configuration manifold mapping to a task space $\mathcal{W} \subset \mathbb{R}^m$, representing the coordinates of the $m\leq 6$ generalized forces acting on the mechanical system. 
The mapping from rotor speeds to the task-space wrench is structurally a nonlinear direct kinematic map between manifolds $f: \mathcal{V} \to \mathcal{W}$, conceptually \emph{analogous to the forward kinematics} of a robotic manipulator. This is \emph{distinct from the linear maps between cotangent spaces relating joint torques to end-effector wrenches in static} manipulator analysis. Assuming fixed-pitch propellers, the lift and drag intensities are proportional to the signed square of the rotational speed. The map $f:\mathcal{V}\to\mathcal{W}$ is formulated as:
\begin{equation}
    v\mapsto f(v) = \sum_{i=1}^n \mathbf{a}_i \operatorname{sgn}(v_i) v_i^2 = \sum_{i=1}^n \mathbf{a}_i v_i |v_i|,
    \label{eq:aerodynamic-map}
\end{equation}
where $\mathbf{a}_i \in \mathbb{R}^m$ represents the $i$-th column of a constant geometric configuration matrix $A$, encoding the generalized forces generated on the mechanical system by the $i$-th rotor for a unitary squared speed~\cite{Hamandi2021IJRR}.

\smallskip
\subsubsection*{The Energy Cost Field}

Given the operational speeds and the low inertia of multirotor blades, the electrical power drawn during flight is  dominated by the dissipation required to overcome aerodynamic drag torque, which satisfies $\tau_{d,i} \propto v_i^2$, yielding a mechanical power consumption $P_{\text{mech}, i} = \tau_{d,i} |v_i| \propto |v_i|^3$. 
We instantiate the independent scalar cost field $h(v)$ introduced in Section \ref{sec:moo_fibers} as the total aerodynamic power cost, modeled by the weighted $L_3$ norm of the actuator speeds:
\begin{equation}
    J_1(v) = \sum_{i=1}^n c_i |v_i|^3,
\end{equation}
with $c_i$ aggregating the  aerodynamic and motor constants~\cite{Bristeau2009ECC}.

\smallskip
\subsubsection*{Fiber Density: Aerodynamic Promptness / Manipulability}

The aerodynamic promptness of the multirotor---defined as its capacity to rapidly alter the body generalized force $w$---is fundamentally bottlenecked by the acceleration bounds of the motors~\cite{Bicego2020JIRS}. This physical limitation maps seamlessly into the abstract geometric framework established in Section \ref{sec:fiber_optimization}.

Considering, as a starting step, the Euclidean Riemannian metric $g$ on the configuration space $\mathcal{V}$ (having $G_v = I_n$, the identity matrix) to represent uniform, normalized constraints on the rotor accelerations ($\|\dot{v}\| \leq 1$), the instantaneous rate of change of the generalized force is entirely governed by the state-dependent Jacobian of the direct kinematic map:
\begin{equation}
    J_f(v) = \tfrac{\partial f}{\partial v} = 2 A \operatorname{diag}(|v|),
\end{equation}
where $\operatorname{diag}(\cdot)$ denotes the diagonal matrix formed by the entries of its array argument.
The pushforward co-metric defined by~\eqref{eq:pushed_forward_metric} simplifies to $D(v) = J_f(v) J_f^T(v)$, thus resulting in:
\begin{equation}
\label{eq:aero-co-metric}
    D(v) = 4 A \operatorname{diag}(v^2) A^T,
\end{equation}
where $v^2 \in \mathbb{R}^n$ is defined as the component-wise square ($(v^2)_i = v_i^2$, for $i=1,\ldots, n$).
The orthogonal fiber density $\rho(v)$, which quantifies a bounded volume of achievable generalized force rates in the task space, evaluates directly to:
\begin{equation}
    v\mapsto\rho(v)  = 2^m \sqrt{\det\big(A \operatorname{diag}(v^2) A^T\big)}.
    \label{eq:aerodyn_promptness}
\end{equation} 
\begin{definition}[Aerodynamic Promptness]
    We designate $\rho:\mathcal{V}\to\mathbb{R}_{\geq 0}$ given by~\eqref{eq:aerodyn_promptness}--i.e.,  the specific physical instantiation  of the fiber density for the aerodynamic map $f$ given by~\eqref{eq:aerodynamic-map}, evaluated with the Euclidean metric on the rotational speed space $\mathcal{V}$-- as the \emph{aerodynamic promptness} or the \emph{aerodynamic manipulability} of the multirotor.
\end{definition}

To integrate this promptness into the multi-objective optimization framework alongside the energetic cost, we define, according to the general theory developed previously:
\begin{equation}
    J_2(v) = - \det\big(A \operatorname{diag}(v^2) A^T\big).
\end{equation}

\smallskip
\subsubsection*{Feasibility on the Multirotor Fiber}

Finally, physical propeller-motor pairs cannot spin infinitely fast, and some are not designed to reverse direction. These and other physical hardware limitations impose, e.g.,  strict upper and lower bounds $v_{\text{min}} \le v_i \le v_{\text{max}}$, formally defining more in general the bounded feasibility region $\overline{\mathcal{V}} \subset \mathcal{V}$. 

\subsection{Future Research: Interplay Between Energy Cost and Aerodynamic Promptness}
\label{subsec:agenda}
The operational multi-objective redundancy resolution for the multirotor thus reduces to evaluating the restricted gradients $\nabla_{\mathcal{F}_w}J_1(v)$ and $\nabla_{\mathcal{F}_w}J_2(v)$ along the truncated feasible fiber $\tilde{\mathcal{F}}_w = \mathcal{F}_w \cap \overline{\mathcal{V}}$. 

We encourage future research to analyze how the interaction between energy and aerodynamic promptness shifts topologies across different base task generalized forces $w$, considering various multirotor designs encoded by the matrix $A$ and their respective physical hardware constraints, all under the umbrella of the proposed framework.

%%% %%%%%%%%%%%%%%%%%%%%%%
\section{Prototypical Case-Study:\\ The Dual-Rotor Actuator}
\label{sec:case_study}
%%% %%%%%%%%%%%%%%%%%%%%%%

\begin{figure}[t]
  \begin{subfigure}[t]{0.49\columnwidth}
    \centering
    \includegraphics[width=\linewidth]{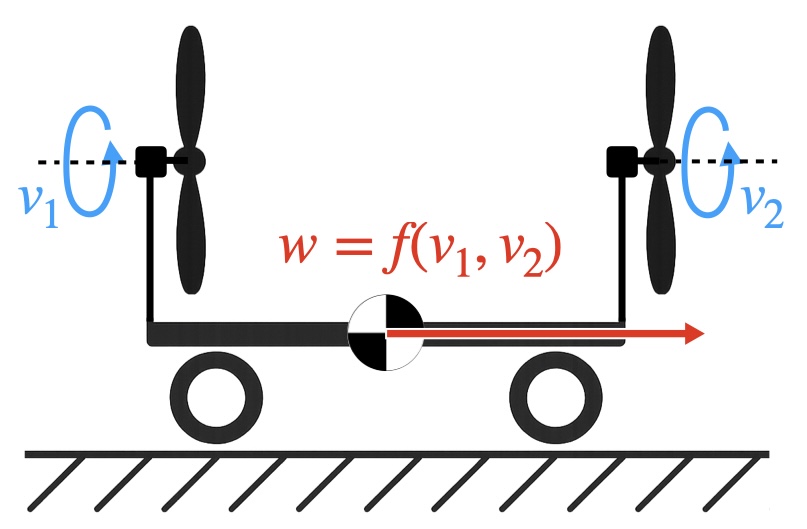}
    \caption{Idealized schematic of a dual-rotor actuator. The output force $w$ applied to the one-dimension-restricted object (a cart) is given by the sum of the two rotor thrusts, which are nonlinearly related to the rotor speeds $v_1$ and $v_2$.}
    \label{fig:dual_rotor_concept}
  \end{subfigure}\hfill
  \begin{subfigure}[t]{0.49\columnwidth}
    \centering
    \includegraphics[width=\linewidth]{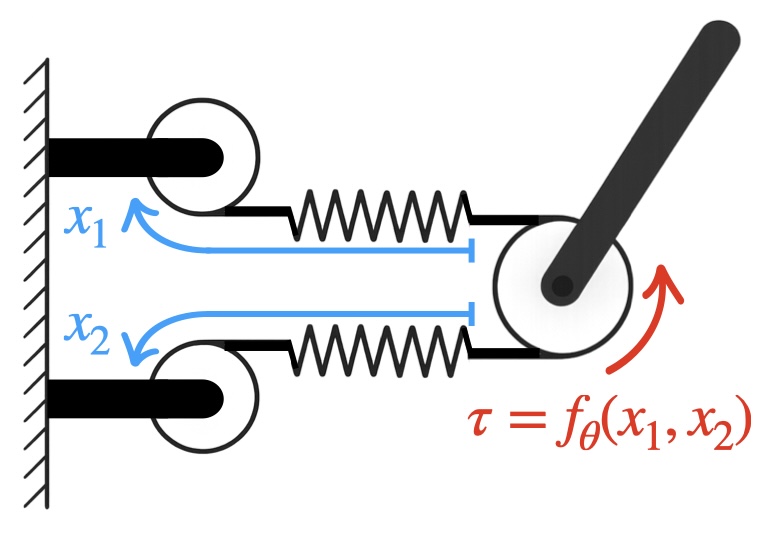}
    \caption{Variable Stiffness Actuator: the output torque $\tau$ is nonlinearly related to  the motor displacements acting on the lengths ($x_1,x_2$) of nonlinear hardening springs (tendons) . The  joint angle $\theta$ parametrizes the nonlinear map.}
    \label{fig:vsa_schematic}
  \end{subfigure}
\caption{A dual-rotor actuated body and a variable stiffness actuator driven robotic link share a structural isomorphism in 1) the nonlinear actuation maps linking the rotor speeds (resp. tendons lengths) to the total force (resp. output torque), and 2) the interplay between the fiber density (aerodynamic manipulability vs torque promptness/variable stiffness) and the energy/metabolic cost (dissipated drag vs dissipated motor current) fields.\vspace{-1em}}
\end{figure}

As a preliminary stepping stone for the proposed research agenda of Section~\ref{subsec:agenda}, we analyze here the simplest redundant system: two rotors controlling a single scalar task variable ($n=2, m=1$), as conceptually illustrated in Fig.~\ref{fig:dual_rotor_concept}.

Let $A = [a_1, a_2]$ be the constant configuration matrix with generic gains $a_1, a_2 \in \mathbb{R} \setminus \{0\}$. The direct kinematic map $f: \mathcal{V} \to \mathcal{W}$ reduces to:
\begin{equation}
    v\mapsto f(v) \coloneqq a_1 \operatorname{sgn}(v_1) v_1^2 + a_2 \operatorname{sgn}(v_2) v_2^2 \;=\; w.
\end{equation}
Applying the formalisms established in the previous section, the competing objectives take a highly transparent form:
\begin{itemize}
    \item \emph{Energy cost ($J_1$):} $J_1(v) = c_1 |v_1|^3 + c_2 |v_2|^3$, with aerodynamic constants $c_i > 0$.
    \item \emph{Aerodynamic Promptness cost ($J_2$):} The aerodynamic manipulability evaluates to $\det(A \operatorname{diag}(v^2) A^T) = a_1^2 v_1^2 + a_2^2 v_2^2$. Thus, the objective is exactly:
    \begin{equation}
        J_2(v) = - (a_1^2 v_1^2 + a_2^2 v_2^2).
    \end{equation}
\end{itemize}
Minimizing $J_2$ pushes the state away from the origin, directly maximizing the weighted sum of the squared rotor speeds to improve control authority. 
Contour plots of the two costs for some specific parametric values (see caption) are shown in Figs.~\ref{fig:dual_A} and~\ref{fig:dual_B}.

\smallskip
\subsubsection{Geometry of the Task Fibers}

\begin{figure*}[t]
  \centering
  % Row 1
  \begin{subfigure}[t]{0.3\textwidth}
    \centering
    \includegraphics[width=\linewidth]{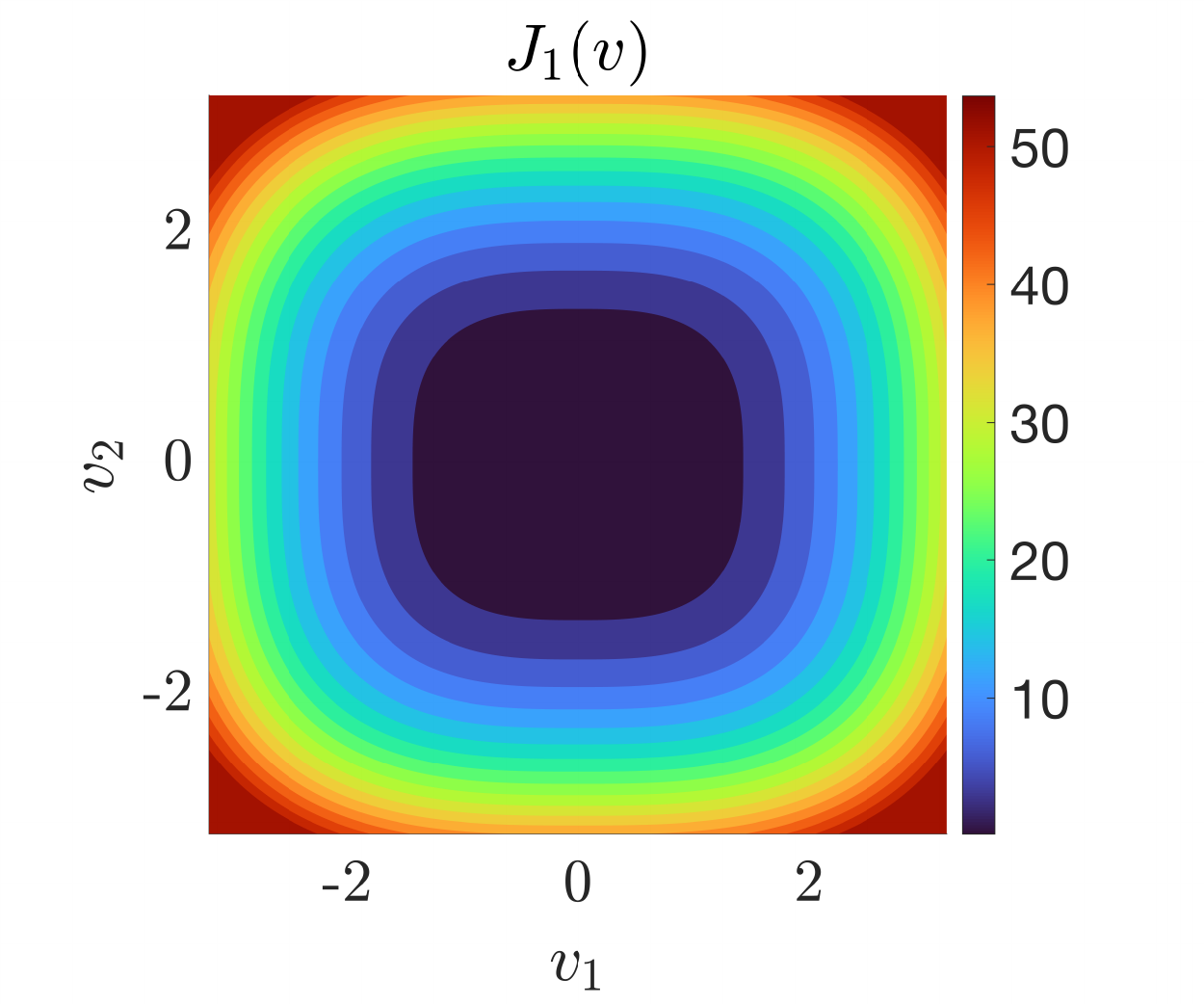}
    \caption{Level sets of the energy cost $J_1$ on $\mathcal{V}$.}
    \label{fig:dual_A}
  \end{subfigure}\hfill
  \begin{subfigure}[t]{0.3\textwidth}
    \centering
    \includegraphics[width=\linewidth]{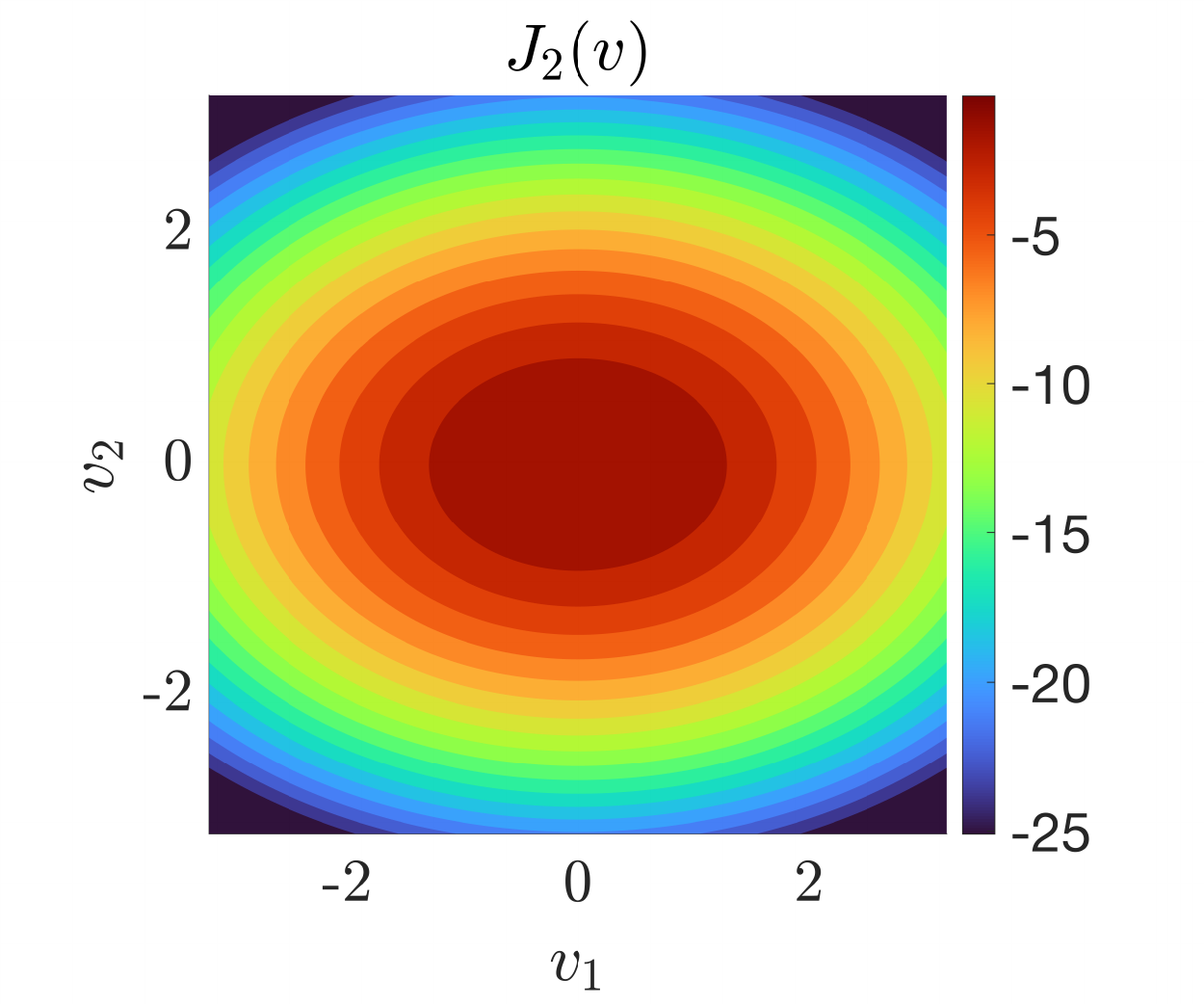}
    % AF: swap the figure to $J_2$ keep the colors.
    \caption{Aerodynamic promptness cost $J_2$ on $\mathcal{V}$.}
    \label{fig:dual_B}
  \end{subfigure}\hfill
  \begin{subfigure}[t]{0.3\textwidth}
    \centering
    \includegraphics[width=\linewidth]{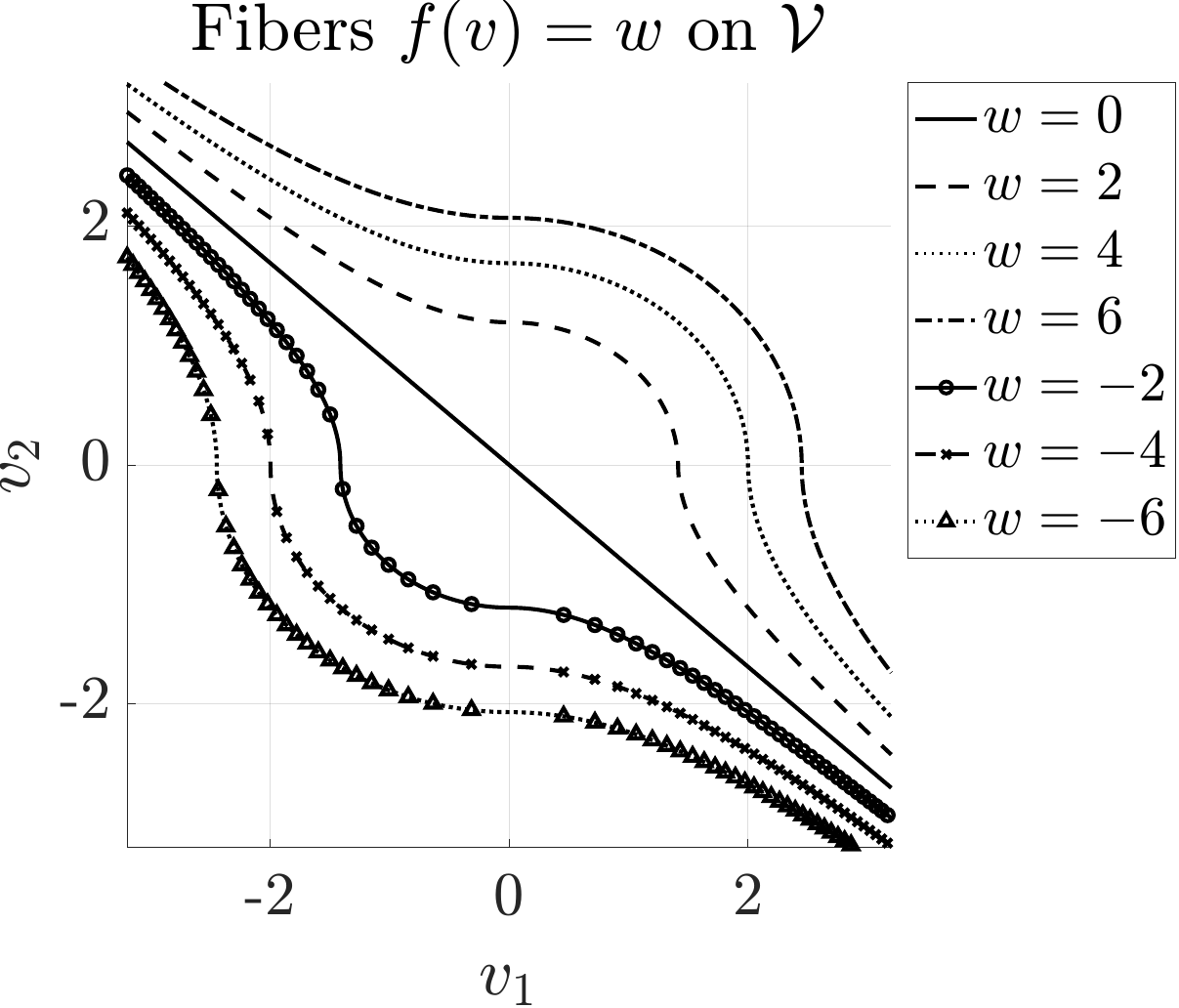}
    \caption{Fibers $\mathcal{F}_w$ on $\mathcal{V}$. 
    }
    \label{fig:dual_C}
  \end{subfigure}

  \caption{Dual-actuator case study summary. Panels (a) and (b) show the competing objective fields. In (c), the task fibers transform from compact to non-compact sets depending on the alignment of the force contributions. Parameters: $a_1=1.0$, $a_2=1.4$, , $c_1=1.0$, , $c_2=1.2$.}
  \label{fig:dual_actuator_panels}
  \vspace{-1em}
\end{figure*}

The topological structure of the  fiber $\mathcal{F}_w \coloneqq \{ v \in \mathbb{R}^2 \mid f(v) = w \}$ is dictated by the signs of the individual generalized force contributions $w_i = a_i \operatorname{sgn}(v_i) v_i^2$. We identify two fundamentally distinct operational regimes:

\begin{enumerate}
    \item \emph{Cooperative actuator region (fibers as compact ellipses):} This occurs when both actuators contribute to the task in the same direction, meaning $\operatorname{sgn}(w_1) = \operatorname{sgn}(w_2) = \operatorname{sgn}(w)$. The fiber equation simplifies to $|a_1| v_1^2 + |a_2| v_2^2 = |w|$. Geometrically, this defines a bounded, compact elliptical arc strictly confined within a single quadrant of $\mathcal{V}$. Because the set is compact, both $J_1$ and $J_2$ are strictly bounded.

    \item \emph{Antagonistic actuator region (fibers as non-compact hyperbolas):} This occurs when the actuators oppose each other, meaning $\operatorname{sgn}(w_1) \neq \operatorname{sgn}(w_2)$. The fiber equation forms an indefinite quadric, such as $|a_1| v_1^2 - |a_2| v_2^2 = w$. Geometrically, these are unbounded hyperbolic branches. The task $w$ is maintained constant by simultaneously increasing both $|v_1|$ and $|v_2|$, representing internal loading or \emph{co-contraction/co-activation}. 
\end{enumerate}
Figure~\ref{fig:dual_C} shows the differing shapes of the fibers in the two regimes. For the particular signs of $a_1>0$ and $a_2>0$ chosen in the example figure, the cooperative regions are the first and third quadrants, and the antagonistic regions are the second and fourth quadrants. 

\begin{figure}[t]
  \centering
  
  \begin{subfigure}[t]{0.49\columnwidth}
    \centering
    \includegraphics[width=\linewidth]{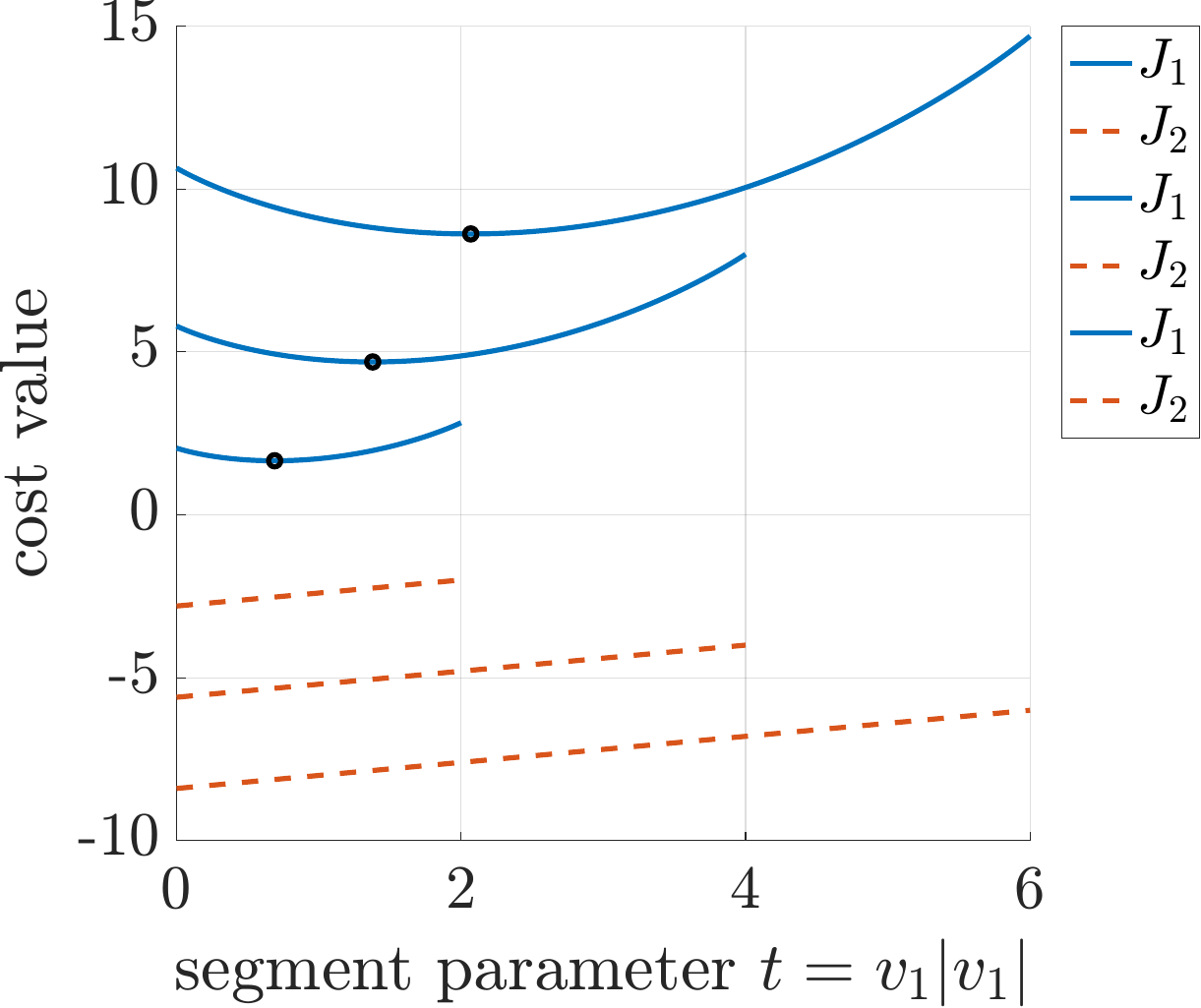}
    \caption{Cooperative region cost profiles along elliptical fibers. $J_1$ attains a unique interior minimum, while $J_2$ is minimized at the boundaries.}
    \label{fig:dual_D}
  \end{subfigure}\hfill
  \begin{subfigure}[t]{0.49\columnwidth}
    \centering
    \includegraphics[width=\linewidth]{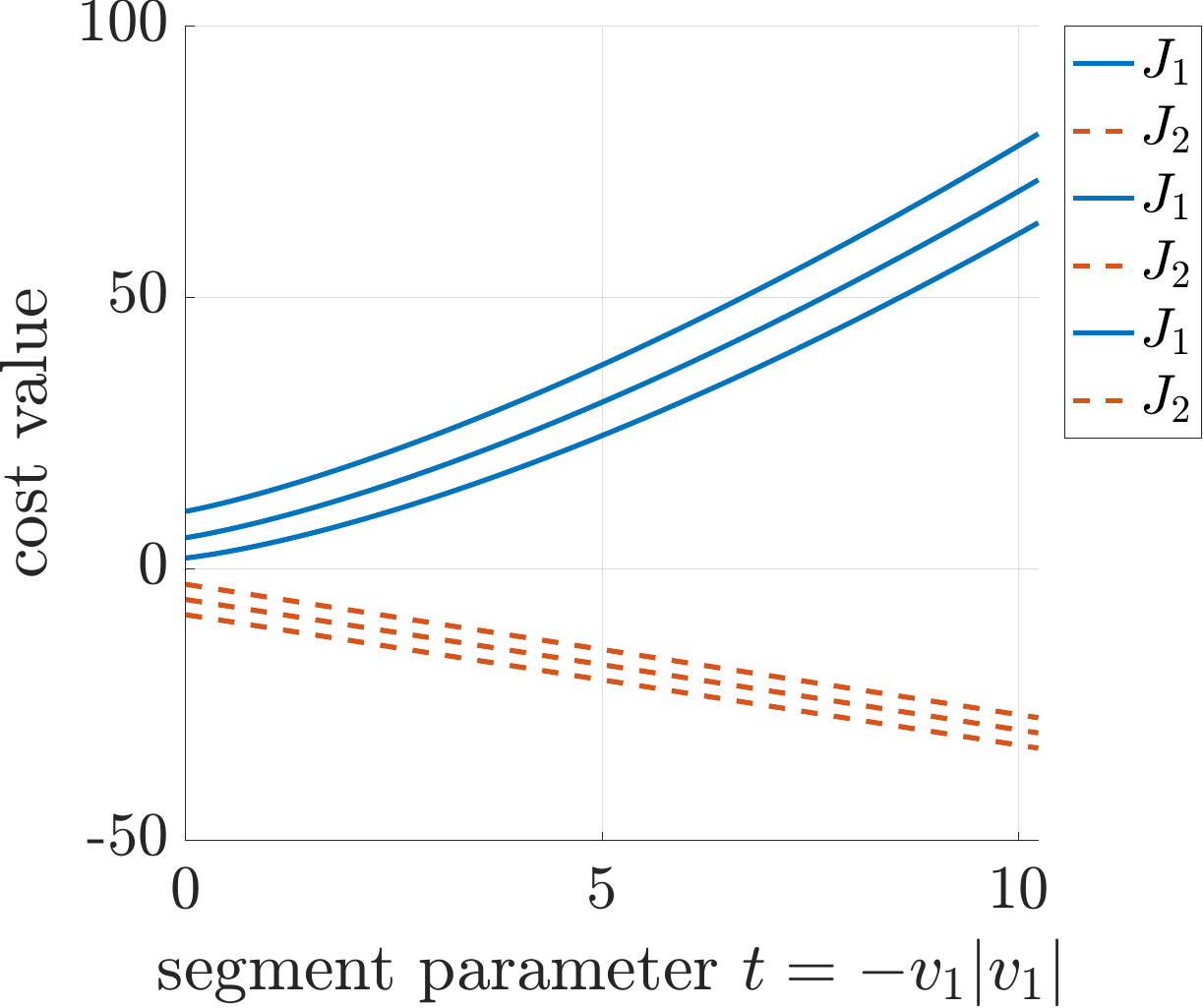}
    \caption{Antagonistic region cost profiles along hyperbolic branches: $J_1$ strictly increases with internal loading, while $J_2$ strictly decreases.}
    \label{fig:dual_E}
  \end{subfigure}

  \caption{Costs $J_1$ and $J_2$ in the cooperative  ($v_1,v_2\geq0$)  and antagonistic ($v_1\geq 0,v_2\geq0$) regions and  evaluated on the fibers corresponding to $w\in\{2,4,6\}$. The topology of the fiber dictates whether the objectives exhibit mild local compatibility or global conflict. Same parameters as Fig.~\ref{fig:dual_actuator_panels}.}
  \label{fig:dual_actuator_profiles}
  \vspace{-1em}
\end{figure}

\smallskip
\subsubsection{Analysis of the Energy Cost and Aerodynamic Promptness Trade-Off}

By evaluating the restricted gradients $\nabla_{\mathcal{F}_w}J_1(v)$ and $\nabla_{\mathcal{F}_w}J_2(v)$ along these distinct topological structures, we can assess the local alignment index $\kappa(v)$.

\subsubsection*{Compatible Trade-Off in the Cooperative Actuator Region}
Along the compact elliptical fibers, energy and promptness exhibit a mild, bounded trade-off. The energy cost $J_1(v)$, driven by its cubic growth, possesses a strict convex basin along the arc that favors an interior load distribution (sharing the effort between both rotors). Conversely, $J_2(v)$ decreases monotonically as the state moves away from the origin, pushing the optimal solution toward the boundary of the quadrant (utilizing only the single actuator with the larger structural gain $|a_i|$). 
\emph{Conclusion:} Compatible trade-off. While the objectives do not share a fiber-stationary point, the Pareto set $\mathcal{P}_w$ is tightly bounded within the finite limits of the elliptical arc. Energy-centric allocation naturally distributes the load without driving promptness to extremely poor values along the fiber. See the plots in Fig.~\ref{fig:dual_D} for a visual intuition.

\subsubsection*{Conflicting Trade-Off in the Antagonistic Actuator Region}
Along the hyperbolic fibers, the relationship fractures into strict conflict. Moving outward along the branch away from the vertex (the point of minimum viable effort), the magnitude of both rotor speeds strictly increases. Consequently, $J_1(v)$ increases monotonically toward infinity, pulling the system back toward the vertex to conserve energy. Simultaneously, the internal loading strictly increases the squared norm of $v$, driving $J_2(v)$ monotonically toward negative infinity. 
\emph{Conclusion:} Total conflict. Evaluating the alignment index consistently yields $\kappa(v) = -1$ everywhere along the branch. Energy optimization demands one of the two forces to be zero, thus having zero internal force, whereas promptness optimization demands as large a co-contraction as possible. The Pareto set $\mathcal{P}_w$ stretches infinitely along the hyperbolic curve, exposing the platform's performance to how the hardware feasibility bounds $\overline{\mathcal{V}}$ ultimately truncate this unbounded conflict. See the plots in Fig.~\ref{fig:dual_E} for a visual intuition.

\smallskip
\subsubsection*{Summary}
This dual-actuator study physically corroborates the geometric hypotheses established in Section~\ref{sec:framework}:
\begin{itemize}
    \item The degree of conflict is not absolute, but critically fiber-dependent. Compact cooperative fibers align the opposing gradients closely enough to yield bounded, manageable Pareto sets.
    \item Non-compact antagonistic fibers perfectly align the restricted gradients in strict opposition ($\kappa = -1$), inducing a fundamental and potentially unbounded dichotomy between energy and aerodynamic promptness/manipulability.
\end{itemize}

%%% %%%%%%%%%%%%%%%%%%
\section{Connection to Muscle-Tendon Models: Variable Stiffness and Fiber Density/Promptness}
\label{sec:vsa_connection}
%%% %%%%%%%%%%%%%%%%%%

The multi-objective conflict exposed along the unbounded hyperbolic fibers in the antagonistic multirotor regime is not an isolated robotic anomaly. Rather, it represents a fundamental physical trade-off that biology has explicitly exploited. To contextualize this, we demonstrate that the mathematical architecture of the simple redundant multirotor described in Sec.~\ref{sec:case_study} is structurally isomorphic to the classic biological muscle-tendon model, artificially realized in robotics as the antagonistic Variable Stiffness Actuator (VSA)~\cite{Wolf-vsa-2016}.

\smallskip
\subsubsection{VSA Kinematics and the Active Torque Map}

Consider a joint actuated by an antagonistic mechanism driven by two independent, high-bandwidth motors, depicted in Fig.~\ref{fig:vsa_schematic}. The motors control the linear displacements $(x_1, x_2)\in\mathcal{V}$ of two tendons routed around an output pulley of radius $R$. 
The tendons incorporate strictly nonlinear, hardening elasticity, modeled by a strictly convex force-displacement function $r(\Delta L)$ where $r' > 0$ and $r'' > 0$. Let $\theta$ be the pulley angle. The net elastic torque exerted on the joint is:
\begin{equation}
    \tau(x_1, x_2, \theta) = R \big[ r(x_1 - R\theta) - r(x_2 + R\theta) \big].
\end{equation}
To exactly parallel the multirotor's direct kinematic map, we isolate the VSA's active force-generation capability and evaluate the required external counterbalancing torque to hold the joint isometric at the zero position ($\theta = 0$). This defines the active torque map $f_{\theta=0}: \mathcal{V} \to \mathcal{W}$:
\begin{equation}
    f_{\theta=0}(x_1, x_2) = R \big[ r(x_2) - r(x_1) \big].
\end{equation}

If we specialize the tendons to purely quadratic hardening springs, $r(\Delta L) = k \Delta L^2$, and denote simply $f_{\theta=0}=f$, the active torque map  reduces to:
\begin{equation}
    x \mapsto f(x_1, x_2) = kR \big( x_2^2 - x_1^2 \big).
\end{equation}
This map is structurally identical to the \emph{antagonistic region} dual-rotor kinematic map $f(v) = -|a_1|v_1^2 + |a_2|v_2^2$ established in Section \ref{sec:case_study}. Just as the multirotor commands a specific task wrench $w$ via an infinite set of rotor speeds lying on a hyperbolic fiber, the VSA sustains a constant isometric load via an infinite set of motor displacements (co-contractions) lying on an identical hyperbolic curve in $\mathcal{V}$.

\smallskip
\subsubsection{Dualism: Passive Stiffness vs. Active Promptness}

In classic VSA literature, traversing the fiber $f^{-1}(\tau)$ is utilized to modulate the joint's passive compliance. The apparent joint stiffness $\sigma$ is the passive resistance to an external angular disturbance:
\begin{equation}
    \sigma(x_1, x_2) = -\left.\frac{\partial \tau}{\partial \theta}\right|_{\theta=0} = R^2 \big[ r'(x_1) + r'(x_2) \big].
    \label{eq:vsa_stiffness}
\end{equation}

However, applying the geometric framework from Sec.~\ref{sec:fiber_optimization}, traversing this exact same fiber dictates the active control authority of the motors through the change of fiber density. The torque promptness/manipulability $\rho$ evaluates the sensitivity of the output torque to instantaneous changes in motor positions:
\begin{equation}
    \rho(x_1, x_2) = \sqrt{\det\big(J_f J_f^T\big)} = R \sqrt{r'(x_1)^2 + r'(x_2)^2}.
    \label{eq:vsa_promptness}
\end{equation}

Physically, $\sigma$ and $\rho$ govern entirely different phenomena: stiffness  maps environmental angular deflections to passive restoring forces ($\Delta \theta \to \Delta \tau$), whereas promptness  maps internal control effort to active torque corrections ($\Delta x \to \Delta \tau$). 
Despite this distinction, they are geometrically bound:
\begin{proposition}  For any strictly hardening spring ($r'' > 0$), increasing internal co-contraction (simultaneously increasing $x_1$ and $x_2$) to hold a constant torque $f(x_1, x_2)=\tau$ strictly increases both the sum $r'(x_1) + r'(x_2)$ and the magnitude $\sqrt{r'(x_1)^2 + r'(x_2)^2}$.  Along any feasible operational fiber, passive stiffness ($\sigma$ in \eqref{eq:vsa_stiffness}) and active promptness ($\rho$ in \eqref{eq:vsa_promptness}), are strictly monotonically related. Increasing joint stiffness intrinsically condenses the torque mapping fibers, yielding a hyper-sensitive transmission gain. Consequently, a VSA operating under  co-contraction functions mathematically as a \emph{Variable Promptness Actuator} (VPA).
\end{proposition}

\subsection{The ``Muscle in the Sky'' Paradigm}

This structural isomorphism directly justifies the overarching premise of this work. Biological systems resolve the redundancy of antagonistic muscle pairs by minimizing energy (metabolic cost) during steady-state operations, while deliberately injecting energy into the null-space (i.e., along the fibers of the nonlinear map) via co-contraction when anticipating rapid disturbances or executing explosive maneuvers. 

A multirotor, when evaluated through the lens of aerodynamic promptness/manipulability, shares this exact mathematical geometry. The rotors can act as nonlinear tendons pulling against the rigid body of the UAV. 
\begin{itemize}
    \item \emph{Energy Mimicry:} In the VSA, pushing the state outward along the hyperbolic task fiber requires increased mechanical deformation (potential energy) and sustained motor holding current. In the multirotor, pushing outward along the same hyperbolic fiber requires massive internal aerodynamic loading (kinetic energy) and sustained motor holding against the dissipating drag moment, driving $J_1(v)$ to extremes.
    \item \emph{Promptness Mimicry:} In both systems, this energetically intense co-contraction strictly maximizes the local fiber density ($\rho$). For the VSA, this yields explosive torque authority. For the multirotor, it yields instantaneous, high-bandwidth wrench generation capable of countering severe aerodynamic disturbances and impacts, or generating agile maneuvers.
\end{itemize}

Ultimately, modulating the internal tensions of a redundant multirotor aircraft is mathematically analogous to flexing a biological joint. By formalizing aerodynamic manipulability, we transition multirotor control allocation from a static force-balancing problem to a dynamic, impedance-like regulation strategy---effectively transforming the concept of the aerial platform into a ``flying muscle'' group.

%%% %%%%%%%%%%%%%%%%%%%%%
\section{Conclusion}
\label{sec:conclusion}
%%% %%%%%%%%%%%%%%%%%%%%%

This work introduced a geometric framework for multirotor redundancy resolution, formalizing the topological trade-off between energetic cost and aerodynamic promptness. We demonstrated that this efficiency-authority dichotomy is governed by task fiber geometry: compact fibers yield bounded Pareto fronts, while unbounded, antagonistic fibers enable aerodynamic co-contraction to maximize promptness. This structural isomorphism to biological variable stiffness actuators establishes a dynamic ``flying muscle'' paradigm.

We encourage leveraging this framework for principled, Pareto-based promptness tuning. Future research should analyze operational trade-offs in physical platforms---specifically coplanar collinear hexarotors, and fully actuated heptarotors or tilted-propeller octorotors---to experimentally validate these endurance-promptness envelopes for next-generation agile aerial robots able to operate, e.g., on harsh  wind conditions for civilian inspection and maintenance.

\printbibliography[title={References}]

\end{document}